\documentclass[runningheads]{llncs}
\usepackage[T1]{fontenc}
\usepackage{graphicx}
\usepackage{xcolor}
\usepackage{booktabs}
\usepackage{multirow}
\usepackage{graphicx}
\usepackage{float}
\usepackage{todonotes}
\usepackage{makecell}

\begin{document}
\title{Trustworthy Visual Quality Inspection under Data Scarcity in Manufacturing}
\titlerunning{Trustworthy Visual Quality Inspection}



%
%
\author{Panagiotis Sapoutzoglou\inst{1}\thanks{These authors contributed equally.} \and
Jessy Ribaira\inst{2}$^{\star}$ \and
Martin Kanounnikoff \inst{2} \and
Bas Tijsma\inst{3} \and
Christian Geiß\inst{2,4} \and
Maria Pateraki\inst{1}}
\authorrunning{P. Sapoutzoglou et al.}
%
\institute{National Technical University of Athens, Athens, Greece \and
German Aerospace Center (DLR), Germany \and
Philips, High Tech Campus 52, 5656 AG Eindhoven, The Netherlands\and
University of Bonn, Bonn, Germany \\
\email{\{psapoutzoglou,mpateraki\}@mail.ntua.gr, \\
\{Jessy.Ribaira, Martin.Kanounnikoff, Christian.Geiss\}@dlr.de, partner.Bas.Tijsma@philips.com}}
\maketitle              
\begin{abstract}
Automated visual inspection in manufacturing aims to replace slow and inconsistent manual checks, but its economic value depends on whether its decisions can be trusted enough to automate routine inspection while reserving human expertise for ambiguous cases.
In production-line settings, defective samples are scarce, since the process is optimized to produce good parts, which limits any learning-based inspector trained on real data alone. Compounding this, defect decisions emitted as hard labels with no confidence estimate carry an asymmetric cost: a false reject wastes a good product, while a false accept may increase the risk of undetected defects progressing through the production process. We address both problems by mitigating data scarcity through the generation of synthetic defective samples with a diffusion model, and meeting the need for confidence-aware decisions with a Bayesian classifier that defers ambiguous units to human review rather than misclassifying them. These components are embedded in a staged pipeline of successive, complementary checks. We evaluate how synthetic augmentation affects classification and localization on a test set of real defects, and examine the system's trustworthiness at three points: the decision, the synthetic data, and the pipeline structure. This work-in-progress reports preliminary results suggesting that diffusion-generated defects, combined with uncertainty-aware classification, can lower the cost of reaching a trustworthy, deployable inspection model under data scarcity.
\keywords{anomaly detection  \and synthetic data generation \and uncertainty quantification \and visual quality inspection.}
\end{abstract}
\section{Introduction}

Despite all efforts made in the context of Smart Industry, the final quality check on production lines is still often performed manually, which at industrial scale becomes a bottleneck through labor cost, inspection time, and inter-operator variability~\cite{Czimmermann_2020_review,hutten2024_review}. Automating it is the obvious step, but a difficult one. The value of a production line lies in its high proportion of non-defective units, which makes defects rare, while learning-based methods perform well only near their training distribution.
This scarcity does more than limit accuracy. A model trained on few samples may confidently assign a known label to a defect lying outside its training distribution, with no indication of the model's internal uncertainty, leaving the manufacturers to either trust all automated predictions or introduce additional manual verification.
A key requirement is therefore the ability to distinguish reliable predictions from uncertain ones, enabling selective human intervention.

Learning-based inspection broadly follows two paradigms. Supervised methods learn defect appearance directly from labeled 
examples and can both detect and classify, but labeled defect data, are often limited in manufacturing environments. Unsupervised methods \cite{batzner2024efficientad,roth2022patchcore} sidestep this by training only on defect-free samples and flagging deviations from normality eliminating the need for defect annotations, but typically providing only detection, not categorization. The later is crucial in production, 
since each defect category points to a different upstream cause (e.g. misaligned printing jig, dust contaminated stencil) and a corrective action. 
Our approach addresses the two above issues — scarce defect data and unreliable decisions — while combining the strengths of both paradigms. We study these problems in the context of print-defect inspection on the chassis of electric shavers, across four defect categories observed in production. 
Our contributions are summarized as follows:
\vspace{-1mm}
\begin{itemize}
\item We propose a staged inspection pipeline that combines region-of-interest segmentation, unsupervised anomaly detection, and uncertainty-aware defect classification.
\item  We present an empirical study of diffusion-based synthetic defect generation under data scarcity, evaluating its impact on defect classification and localization using real production defects.
\item  We investigate uncertainty-aware defect classification using Bayesian neural networks and deep ensembles, assessing their predictive performance and calibration in low-data regimes.


\end{itemize}





\section{Related work}

\subsection{Synthetic defect generation}
\vspace{-1mm}
As labeled defects are scarce in manufacturing, a growing body of work generates them synthetically to augment training data \cite{Czimmermann_2020_review}. Early generative approaches were based on GANs \cite{Zhang2021DefectGANHD,Duan2023DFMGAN} that synthesize defects by simulating defacement and restoration over a normal background, and showed that GAN-based generated defects can improve recognition when real samples are limited \cite{niu_2019_defectgan}, including in real production settings \cite{ROZANEC202311094}. More recently, diffusion models have improved the realism and diversity of few-shot anomaly generation. AnomalyDiffusion \cite{hu2023anomalydiffusion} focuses on generating the anomaly region, while DualAnoDiff \cite{jin2024dualanodiff} jointly generates the full image and the corresponding anomaly part through two interrelated diffusion branches, producing pixel-aligned masks and reporting gains on downstream detection, localization, and classification. DualAnoDiff is particularly relevant to our case: a single generated sample yields both an image and a mask, supporting classification and localization at once.

\vspace{-2mm}
\subsection{Defect detection}
\vspace{-1mm}
Learning-based defect detection is commonly categorized into supervised and unsupervised paradigms, reviewed comprehensively in \cite{tao2022survey,liu2024survey}. Supervised methods learn defect appearance from labeled examples and can typically both detect and classify, but they require defect data that is costly to collect and especially to annotate. Unsupervised methods instead model the distribution of defect-free samples and flag deviations from it, sidestepping defect scarcity, since the vast majority of manufactured products are defect-free. On the dedicated anomaly-detection benchmark MVTec AD \cite{bergmann2019mvtec}, embedding-based methods such as PaDiM \cite{defard2021padim} and PatchCore \cite{roth2022patchcore} achieve strong detection and localization by comparing test features against a distribution learned from good samples. EfficientAD \cite{batzner2024efficientad} pushes this paradigm further, reaching the millisecond-level latency and low memory required for in-line deployment while retaining competitive accuracy. Our pipeline uses EfficientAD as the stage that decides whether an anomaly is present, both for its low latency, which suits a production line, and because training only on good samples means its decision is less dependent on the representativeness of the available defect set.

\vspace{-2mm}
\subsection{Uncertainty estimation}
\vspace{-1mm}
Uncertainty is decomposed into aleatoric (observation noise) and epistemic (model lack of knowledge) components \cite{uncertainty_kendall}. In defect inspection, epistemic uncertainty is critical as scarce defect samples leaves the model’s training coverage sparse \cite{Habibpour2021AnUD}. Therefore, classifiers produce overconfident but wrong predictions for unseen classes under distribution shifts \cite{data_shift_ovadia,survey_Gawlikowski}, undermining inspection pipelines.
To tackle this challenge, proposed approaches include Bayesian Neural Networks (BNNs), which place prior distributions over weights \cite{weight_uncertainty_blundell,uncertainty_kendall,survey_Gawlikowski}, or Monte Carlo Dropout (MC Dropout), which approximates a BNN via stochastic forward passes \cite{dropout_gal16} without architectural changes - suiting resource-constrained industrial applications \cite{survey_Gawlikowski}. Alternatively, deep ensembles improve calibration and robustness, but at higher computational and memory costs \cite{data_shift_ovadia,survey_Gawlikowski,deep_ensembles_Lakshminarayanan}.
Prior studies demonstrated that MC dropout uncertainty correlates with misclassifications, enabling targeted human review of uncertain predictions \cite{SAJEDI2021126}, and Bayesian methods reject more effectively out-of-distribution samples than deterministic baselines \cite{Habibpour2021AnUD}. Consequently, these results position uncertainty estimates as a decision variable within broader inspection pipelines, and coupled with synthetic data generation, they can bridge distributional gaps that arise from limited real-world defect data.

\section{Proposed pipeline}
\label{sec:pipeline}

\begin{figure}[t]
    \centering
    \includegraphics[width=0.9\linewidth]{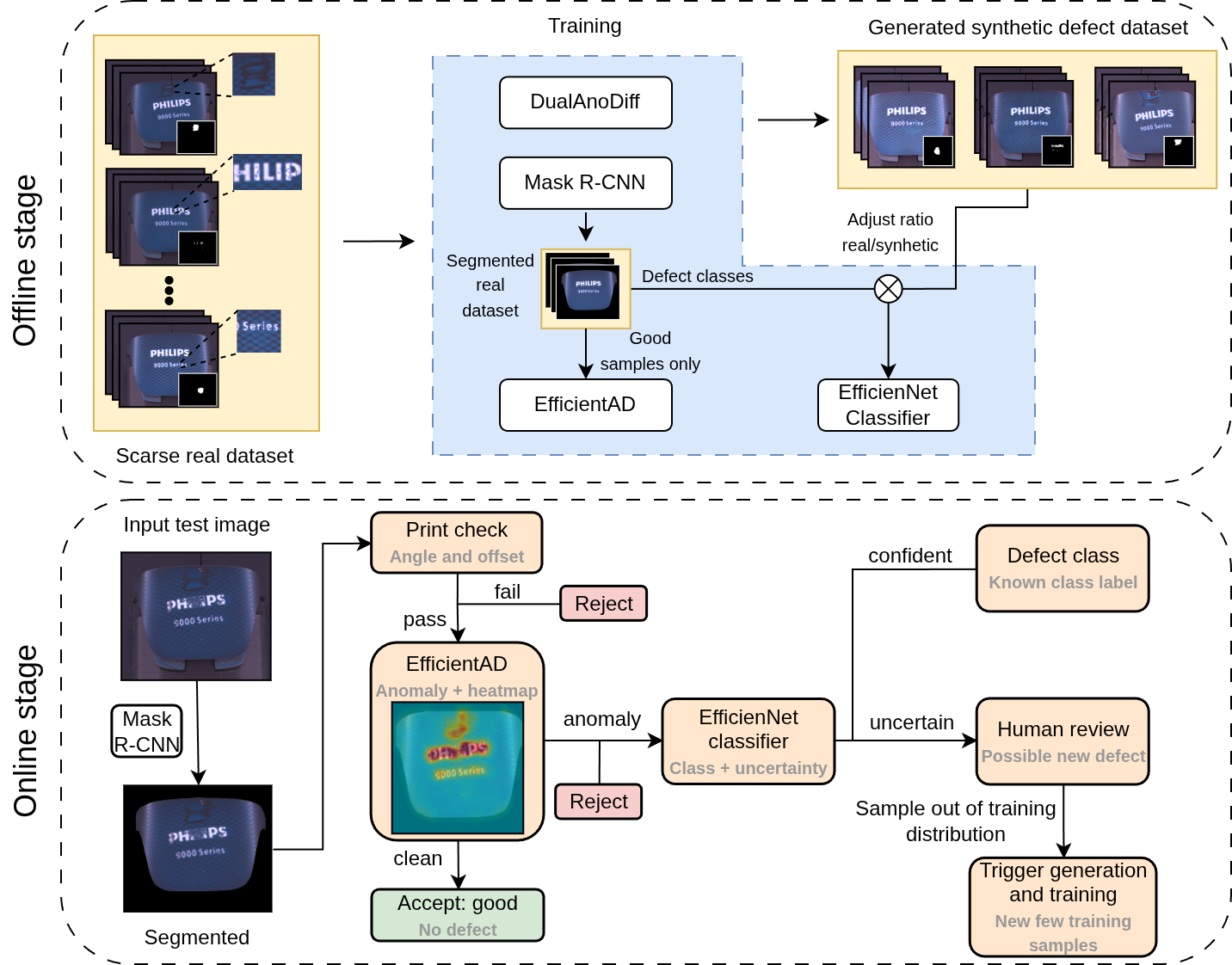}
    \caption{Overview of the proposed pipeline. Offline: DualAnoDiff~\cite{jin2024dualanodiff} generates synthetic defects to train the EfficientNet classifier, while EfficientAD~\cite{batzner2024efficientad} is trained on defect-free samples. Online: a segmented image passes a print check, EfficientAD anomaly detection, and EfficientNet classification, deferring uncertain units to human review.}
    \label{fig:pipeline_overview}
\end{figure}
\vspace{-1mm}
Figure~\ref{fig:pipeline_overview} shows the proposed pipeline, split into two stages: an offline stage, in which synthetic defect images are produced (Sec.~\ref{subsec:syndata_gen}) and the relevant models are trained (Sec.~\ref{subsec:unsupervised_detection} and~\ref{subsec:classification}), and an online stage, which applies a sequence of progressive checks (print check, unsupervised anomaly detection, and defect classification with uncertainty estimation) in a structured inspection of each incoming unit.
As a work in progress, our preliminary evaluation focuses on the offline stage — specifically, whether synthetic defect generation alleviates data scarcity and improves classifier accuracy (Sec.~\ref{sec:results}). The end-to-end online pipeline is left for evaluation in the near future.

\vspace{-2mm}
\subsection{Synthetic data generation}
\label{subsec:syndata_gen}
\begin{figure}[htb]
    \centering
    \includegraphics[width=1\linewidth]{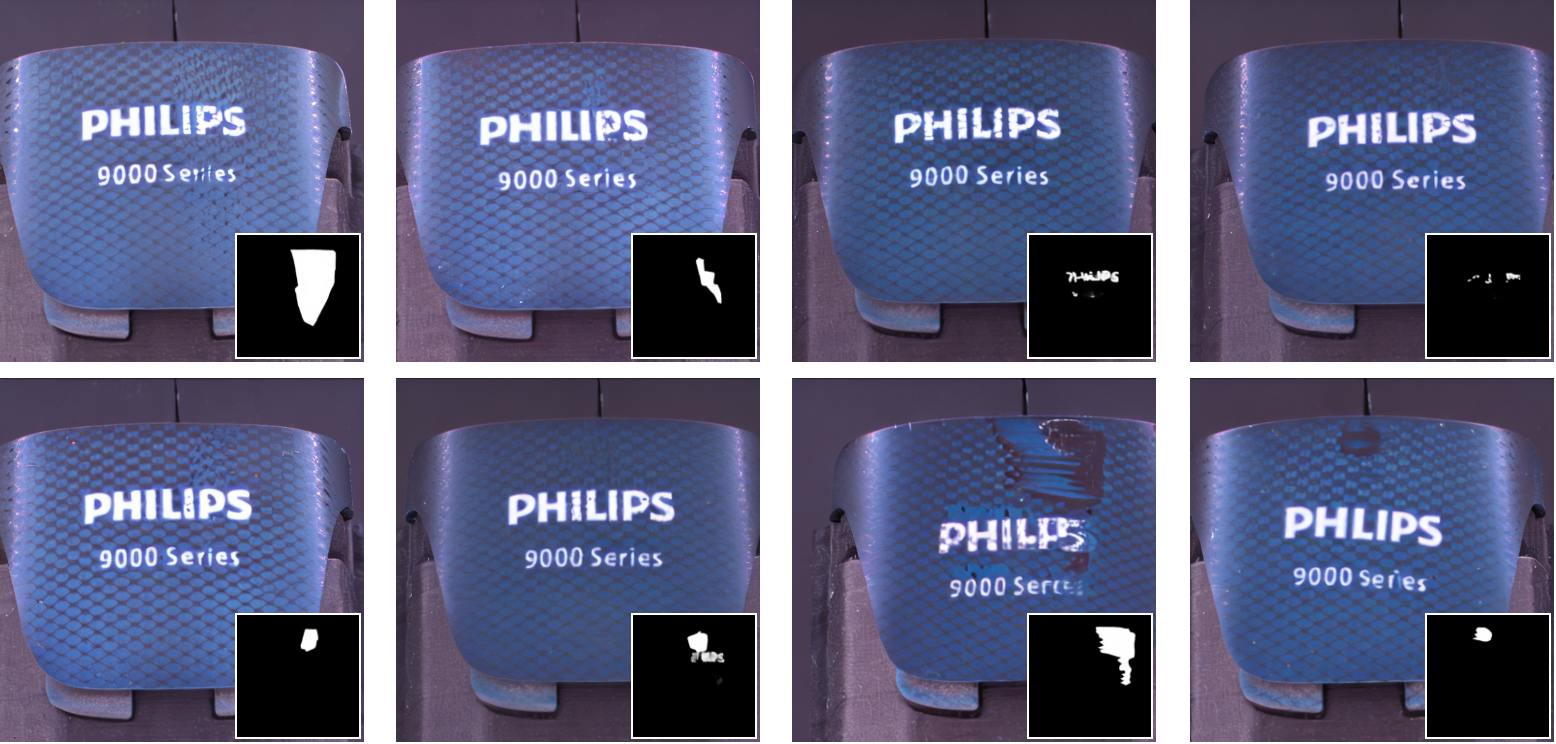}
    \caption{Synthetic defect samples generated with DualAnoDiff with their corresponding generated masks.
    }
    \label{fig:gen_checkpoints}    
\end{figure}

\vspace{-1mm}
To generate synthetic defect samples from a limited initial set of real samples (50 images in total, divided into 10, 30, 10 samples of the fingerprint, interrupted\_print and smudged categories, respectively) we use DualAnoDiff\cite{jin2024dualanodiff}, a dual-branch interrelated diffusion model built on Stable Diffusion v1.5. Rather than inpainting a defect into a fixed background, DualAnoDiff jointly generates two interrelated views in a single denoising process: a global branch that renders the complete anomalous object and a foreground branch that renders the isolated defect region, with the two branches sharing attention so that the synthetized anomaly and its generated mask are consistent.We fine-tune per defect type for 10k steps, saving a checkpoint every 1000 steps. We select the best checkpoint by generating 50 probe images, scored against the training pool by two complementary metrics: IC-LPIPS and KID, and we keep the checkpoint maximizing their average. For each defect we generate 500 images and the corresponding masks (Fig.\ref{fig:gen_checkpoints}). The masks are derived after passing the foreground images through $U^2$-Net~\cite{Qin_2020_PR}.

\vspace{-2mm}
\subsection{Unsupervised anomaly detection}
\label{subsec:unsupervised_detection}
\vspace{-1mm}
The input image is first segmented to isolate the shaver chassis from the background using Mask R-CNN, so that subsequent stages operate only on the region of interest and are not distracted by background variation. The first learning stage assesses whether a unit deviates from normality without relying on the defect types. We adopt EfficientAD \cite{batzner2024efficientad}, motivated by two properties suited to in-line development: it trains exclusively on defect-free samples, and it has low latency. It couples a lightweight student–teacher model, whose student fails to reproduce the teacher's features on unseen (anomalous) regions, with an autoencoder that captures global structure, yielding a combined anomaly map and a scalar anomaly score.
In our pipeline, EfficientAD receives the segmented chassis image and produces both an anomaly score and a pixel-level heatmap. Units scoring below the operating threshold are accepted as defect-free, whereas flagged units are passed to the classification stage that assigns the defect category and the uncertainty. This keeps detection agnostic to the defect taxonomy; it determines whether an anomaly is present and localizes the affected region, leaving the kind of defect to the classifier.

\vspace{-2mm}
\subsection{Defect classification and uncertainty estimation}
\label{subsec:classification}
\vspace{-1mm}
In order to compute both aleatoric and epistemic uncertainty in defect classification, we employed two approaches: deep ensembles and an approximation to a Bayesian neural network.
Firstly, a deep ensemble framework was implemented using the EfficientNet-B3 architecture \cite{Tan2019EfficientNetRM} as the base model, and coupled with a classification head. The ensemble consists of 5 independently trained members, each initialized with a distinct random seed. All models were pre-trained on ImageNet and fine-tuned on our defect datasets. Secondly, the EfficientNet-B3 was modified to approximate a BNN. This model is built on a hierarchy of Mobile Inverted Bottleneck Convolution blocks (MBConv), organized into stages: early layers capturing low-level features, and deeper layers encoding high-level semantics. To introduce Bayesian properties, a Bayesian CNN-style modification \cite{labonte2020knowdontknow3d} was adopted: the early stages (blocks 1-5) were kept deterministic for stable feature extraction, and final stages were adapted by replacing the standard Conv2d layers with Flipout layers \cite{wen2018flipout}. These layers apply pseudo-independent perturbations to the weights during forward passes.

\section{Experiments and results}
\label{sec:results}
\subsection{Setup}
\vspace{-1mm}


\textbf{Dataset}. The dataset consists of Philips electric shavers' chassis images, spanning five classes
: a defect-free class and four defect classes (fingerprint, smudged, interrupted print and print position wrong). Furthermore, synthetic images are generated for fingerprint, smudged and interrupted print classes. The defect classifier and defect generator focus on these three classes, whereas the anomaly detector uses the defect-free class; which was trained on 50 real defect-free images. Due to the scarcity of real samples, the classifier experiments using the Bayesian and Ensemble approaches were conducted on training sets containing 10 real images each for the smudged and fingerprint classes, and 15 real images for the interrupted print class. The remaining samples required for each experimental configuration are synthetically generated. All classification results are evaluated exclusively on an independent test set of the remaining real defect images only, comprising of 24 images in total: 8 per class. For the binary task, the test set therefore consists of 16 real images. To ensure a balanced evaluation across defect categories, the test set size was determined by the smallest available class (smudged), while the same number of samples was selected from the other classes. Finally, the classifier was evaluated under two experimental configurations: a binary classification, trained on the classes fingerprint and smudged, and a three-class classification, trained on the fingerprint, smudged and interrupted print classes.
\newline
\textbf{Evaluation metrics}. The classifiers were evaluated using classification performance metrics (Accuracy, Precision, Recall, F1-score and AUROC), calibration metrics (Brier score for probability accuracy, and Expected Calibration Error (ECE) for confidence calibration), and system metrics (throughput and latency).
\vspace{-2mm}
\subsection{Experimental results}
Table \ref{tab:model_comparison_classifier} compares both classifying approaches on binary and multi-class tasks. On the binary task, both achieved equivalent classification performance on the 16 real test images, yielding an accuracy of 1.000 across all experimental configurations. Given the small size of the test set, this result should be interpreted with caution as a single misclassification would reduce the accuracy. The perfect scores are consistent with fingerprint and smudged being visually well-separated defect types and evaluation on a larger rest test set is left for future work as more data becomes available. Nonetheless, the Bayesian classifier yields better calibration, with Brier and ECE reduced by up to two orders of magnitude, suggesting high model confidence on well-represented classes. On the three-class task, the Bayesian approach outperforms Deep Ensemble in accuracy and F1-score as training data increases (0.833 and 0.830 at 500 images per class), and higher AUROC (0.977 vs 0.930). Beyond performance, epistemic uncertainty \textcolor{blue}{should} provide an actionable inspection signal: low uncertainty flags in-distribution inputs, whereas elevated epistemic uncertainty identifies likely out-of-distribution samples (ie. unseen defect classes) for human review. This supports a human-in-the-loop pipeline that remains robust to novel defect classes. For the binary task, epistemic uncertainty is reported as 0.00 across all Bayesian experiment, consistent with the small test set. The Bayesian model's stochastic forward passes agree with high consistency on every image, driving the variance-based epistemic estimation to zero. In contrast, the Deep Ensemble reports non-zero epistemic uncertainty on the same task as its estimate reflects the disagreement across independently trained models rather than a single one. This indicates that the Bayesian models zero values likely reflect the estimation method's sensitivity to single model confidence on a small test set, rather than an absence of model uncertainty. Finally, regarding the aleatoric uncertainty, it decreases with more training data, in the binary task, which may suggest that the inherent class ambiguity becomes better calibrated with increased data. For the three class task, aleatoric uncertainty initially decreases but rises at 500 images. As more synthetic samples are added across the three classes, the likelihood of ambiguous or borderline examples between the classes grows and greater is their overlap among themselves.

Moreover, the measured latency and throughput from Table \ref{tab:performance_comparison_avg} are consistent with the computational characteristics of each uncertainty estimation approach. Deep Ensemble provides lower latency and higher throughput, as inference requires only forward passes through multiple deterministic models followed by prediction aggregation. In contrast, the Bayesian approach introduces additional computational overhead due to Monte Carlo sampling and its architecture. Consequently, the Bayesian model achieves more comprehensive uncertainty estimation at the cost of increased inference time.
Overall, the pipeline's efficient individual steps result in a total latency that makes it well-suited for online applications and production-line deployment.

\begin{table*}[htb]
\centering
\caption{Comparison of Deep Ensemble and Bayesian models under different dataset configurations. The number of images represents the number of images per class.For the calibration metrics and uncertainty, the metric has been multiplied by the given factor in the parenthesis.}
\label{tab:model_comparison_classifier}
{
\resizebox{\textwidth}{!}{
\begin{tabular}{lc|c|c|c|c|c|c|c|c|c|c|c}
\toprule
  & \multicolumn{6}{c}{Deep Ensemble} & \multicolumn{6}{c}{Bayesian} \\
\cmidrule(lr){2-7}\cmidrule(lr){8-13}
& \multicolumn{3}{c}{2 Classes} & \multicolumn{3}{c}{3 Classes} & \multicolumn{3}{c}{2 Classes} & \multicolumn{3}{c}{3 Classes} \\
\cmidrule(lr){2-7}\cmidrule(lr){8-13}

\makecell{Number of \\images per class} & 50 & 100 & 500 &  50 & 100 & 500 & 50 & 100 & 500 & 50 & 100 & 500 \\
\midrule
\multicolumn{13}{c}{\textbf{Performance metrics}} \\
\midrule
Accuracy   & 1.000 & 1.000 & 1.000 & 0.625 & 0.750 & 0.750 & 1.000 & 1.000 & 1.000 & 0.667 & 0.750 & \textbf{0.833}\\
Precision  & 1.000 & 1.000 & 1.000 & 0.500 & 0.857 & \textbf{0.857} & 1.000 & 1.000 & 1.000 & 0.500 & 0.778 & 0.853\\
Recall     & 1.000 & 1.000 & 1.000 & 0.625 & 0.750 & 0.750 & 1.000 & 1.000 & 1.000 & 0.667 & 0.750 & \textbf{0.833}\\
F1         & 1.000 & 1.000 & 1.000 & 0.533 & 0.709 & 0.709 & 1.000 & 1.000 & 1.000 & 0.556 & 0.733 & \textbf{0.830}\\
AUROC      & 1.000 & 1.000 & 1.000 & 0.841 & 0.924 & 0.930 & 1.000 & 1.000 & 1.000 & 0.935 & 0.903 & \textbf{0.977}\\
\midrule
\multicolumn{13}{c}{\textbf{Calibration metrics and Uncertainty}} \\
\midrule
Brier ($\times10^{3}$)      & 5.00 & 1.30 & 4.40 & 188.0 & 111.0 & 110.0 &
                               0.019 & 0.002 & \textbf{0.001} &
                               135.0 & 107.0 & 94.5 \\

ECE ($\times10^{2}$)        & 6.00 & 2.56 & 3.74 & 27.3 & 10.9 & 19.7 &
                               0.232 & 0.108 & \textbf{0.048} &
                               23.7 & 9.20 & 24.5 \\

Epistemic ($\times10^{2}$)  & 1.24 & 3.20 & 7.03 & 1.69 & 7.96 & 2.75 &
                               0.00 & 0.00 & 0.00 &
                               4.03 & 0.12 & 0.161 \\

Aleatoric ($\times10$)  & 2.00 & 0.75 & 0.506 & 3.06 & 2.67 & 3.93 &
                               0.755 & 0.409 & 0.189 &
                               1.16 & 8.55 & 2.25 \\
\bottomrule
\end{tabular}}}
\end{table*}

\begin{table*}[!t]
\centering
{
\caption{Inference system performance based on the different steps (chassis segmentation, print check, anomaly detection with EfficientAD, and classifier) of the pipeline from Fig \ref{fig:pipeline_overview} .}
\label{tab:performance_comparison_avg}

\begin{tabular}{lcc}
\toprule
& \textbf{Throughput (img/s)  } & \textbf{Latency (s)} \\

\midrule

Segmentation & 5.26 & 0.1911 \\
Print Check  & 25.96  & 0.0391  \\
Anomaly & 3.99 & 0.2503  \\

\midrule
\multicolumn{3}{c}{Classification} \\
\midrule

Bayesian & 1.97 & 0.5073  \\
Deep Ensemble & 15.90 & 0.0629  \\

\midrule
\multicolumn{3}{c}{Total pipeline} \\
\midrule

Bayesian & 1.00 & 0.9985 \\
Deep Ensemble & 1.88 & 0.5324 \\

\bottomrule
\end{tabular}%
}
\end{table*}

\vspace{-2mm}
\section{Conclusion and future work}
\vspace{-1mm}
We presented a work-in-progress inspection pipeline targeting two obstacles to economically viable inspection automation: the scarcity of defect data on optimized production lines, and the cost of acting on untrustworthy decisions. Synthetic generation relieves data scarcity at near-zero marginal cost relative to collecting rare real defects, and our preliminary results suggest synthetic augmentation improves classifier accuracy and calibration. Pairing classification with uncertainty estimation lets the pipeline defer only the decisions it cannot be trusted on to a human, preserving the labor savings of automation. This selective human involvement can additionally exploit the contextual and experiential knowledge of expert operators, which may be particularly valuable for rare or previously unseen cases not adequately represented in the training data.
These results are preliminary and limited to the offline stage. In ongoing work we will evaluate the end-to-end online pipeline and 
close the loop in Fig. \ref{fig:pipeline_overview}, where out-of-distribution units trigger few-shot generation and retraining for new defect types.
\vspace{-2mm}
\begin{credits}
\subsubsection{\ackname}
This work is supported by the Horizon Europe project PANDORA under
Grant Agreement No 101135775. Special thanks to Athanasios Balachtsis for his support on synthetic data generation.
\end{credits}

\bibliographystyle{splncs04etal}
\bibliography{references}

\end{document}